\documentclass{article}

\usepackage[dblblindworkshop,final]{neurips_2025}
\usepackage{textgreek}
\usepackage[utf8]{inputenc} % allow utf-8 input
\usepackage[T1]{fontenc}    % use 8-bit T1 fonts
\usepackage{hyperref}       % hyperlinks
\usepackage{url}            % simple URL typesetting
\usepackage{booktabs}       % professional-quality tables
\usepackage{algorithm}
\usepackage{algpseudocode}   % from algorithmicx
\usepackage{amsfonts}       % blackboard math symbols
\usepackage{nicefrac}       % compact symbols for 1/2, etc.
\usepackage{microtype}      % microtypography
\usepackage{xcolor}         % colors
\usepackage{amsmath}
\usepackage{enumitem}
\usepackage{booktabs,siunitx,caption}
\workshoptitle{The Second Workshop on GenAI for Health: Potential, Trust, and Policy Compliance}
\usepackage{subcaption}
\usepackage{caption}
\usepackage{graphicx}
\usepackage{geometry}
\usepackage{longtable}
\usepackage{multirow}
\usepackage{siunitx}
\usepackage{caption}
\usepackage{wrapfig}
\title{High-Fidelity Synthetic ECG Generation via Mel-Spectrogram Informed Diffusion Training}

\author{
Zhuoyi Huang$^{1}$ \quad Nutan Sahoo$^{1}$ \quad Anamika Kumari$^{1}$ \quad Girish Kumar$^{1}$  \\
\textbf{Kexuan Cai}$^{1}$ \quad \textbf{Shixing Cao}$^{1}$ \quad \textbf{Yue Kang}$^{1}$ \quad \textbf{Tian Xia}$^{1}$  \\ \textbf{Somya Chatterjee}$^{1}$ \quad \textbf{Nicholas Hausman}$^{1}$ \quad \textbf{Aidan Jay}$^{1}$ \quad \textbf{Eric S. Rosenthal}$^{2}$  \\
 \textbf{Soundar Srinivasan}$^{1}$ \quad \textbf{Sadid Hasan}$^{1}$ \quad \textbf{Alex Fedorov}$^{3}$ \quad \textbf{Sulaiman Vesal}$^{1}$ \\
$^1$Microsoft \quad $^2$Massachusetts General Hospital, Harvard University \quad $^3$Emory University
}

\begin{document}

\sisetup{
    detect-mode,
    detect-family,
    table-number-alignment = center,
    round-mode = places,
    round-precision = 4
}

\maketitle
\begin{abstract}
The development of machine learning for cardiac care is severely hampered by privacy restrictions on sharing real patient electrocardiogram (ECG) data. Although generative AI offers a promising solution, the real-world use of existing models-synthesized ECGs is limited by persistent gaps in trustworthiness and clinical utility. In this work, we address two major shortcomings of current generative ECG  methods: insufficient morphological fidelity and the inability to generate personalized, patient-specific physiological signals. To address these gaps, we build on a conditional diffusion-based Structured State Space Model (SSSD-ECG) with two principled innovations: (1) \textbf{MIDT-ECG (Mel-Spectrogram Informed Diffusion Training)}, a novel training paradigm with time-frequency domain supervision to enforce physiological structural realism, and (2) multi-modal demographic conditioning to enable patient-specific synthesis. We comprehensively evaluate our approach on the PTB-XL dataset, assessing the synthesized ECG signals on fidelity, clinical coherence, privacy preservation, and downstream task utility. MIDT-ECG achieves substantial gains: it improves morphological coherence, preserves strong privacy guarantees with all metrics evaluated exceeding the baseline by 4\%-8\%, and notably reduces the interlead correlation error by an average of 74\%, while demographic conditioning enhances signal-to-noise ratio and personalization.  In critical low-data regimes, a classifier trained on datasets supplemented with our synthetic ECGs achieves performance comparable to a classifier trained solely on real data. Together, we demonstrates that ECG synthesizers, trained with the proposed time–frequency structural regularization scheme, can serve as personalized, high-fidelity, privacy-preserving surrogates when real data are scarce, advancing the responsible use of generative AI in healthcare.
\end{abstract}

\section{Introduction}

Cardiovascular disease remains the leading cause of death worldwide, creating a staggering health and economic burden~\citep{mensah2023global}. The electrocardiogram (ECG) is the cornerstone of cardiac diagnostics, and applying machine learning to these signals promises earlier and more accurate diagnoses~\citep{martinez2023current}. However, this promise is constrained by a fundamental data access bottleneck. ECGs are not merely medical records; they are sensitive biometric data that reveal extensive personal health information~\citep{sun2023towards}. Consequently, privacy regulations limit the sharing of large, diverse datasets needed to train robust and generalizable AI models. High-fidelity synthetic data generation has emerged as the most promising solution, offering a pathway to democratize research and accelerate innovation~\citep{delaney2019synthesis,alcaraz2023diffusion}.

High-fidelity synthetic data generation has emerged as the most promising solution to this fundamental challenge~\citep{delaney2019synthesis,alcaraz2022diffusion}. However, the field faces a two-fold technical gap that current state-of-the-art models, such as SSSD-ECG~\citep{alcaraz2023diffusion}, have yet to address fully. First, current models typically condition only on coarse diagnostic labels, producing “one-size-fits-all” signals that ignore patients’ unique demographic variation (e.g., age, sex). This lack of personalization reduces their applicability in real-world research and downstream clinical tasks. The second is the \textbf{morphological fidelity gap}. Prevailing approaches rely heavily on time-domain pointwise losses such as mean squared error (MSE). While simple, these losses fail to enforce global structural properties of ECG waveforms, such as inter-lead correlations and the precise morphology of the P–QRS–T complex. As a result, synthetic signals may achieve low reconstruction error but lack diagnostic realism and trustworthiness.

To address these limitations, we propose two principled enhancements to the state-of-the-art diffusion framework. To bridge the personalization gap, we introduce a multimodal conditioning mechanism that fuses patient demographics with clinical labels, enabling fine-grained, patient-specific generation. To address the morphological fidelity gap, we introduce \textbf{MIDT-ECG (Mel-Spectrogram Informed Diffusion Training for ECGs)}, a novel training paradigm that imposes a rational prior on the signal's time-frequency structure. By emphasizing diagnostically relevant low-frequency bands while capturing multi-scale spectral detail, this loss enforces morphological and physiological plausibility beyond what classic point-wise MSE can achieve. We also include a comprehensive evaluation framework that spans multiple dimensions often overlooked by existing works to validate the high efficiency of our proposed method.

% Our goal is not merely to improve point-wise error but to enhance the physiological plausibility of the generated data, which we prove through a comprehensive evaluation that includes a clear improvement in inter-lead correlation preservation. 
% These contributions collectively establish a robust methodology for synthesizing trustworthy, personalized medical time series, providing a scalable and privacy-preserving foundation for cardiovascular AI research.
Our contributions can be summarized as follows:
\begin{itemize}
    \item We introduce a disentangled \textbf{conditioning framework} that fuses clinical and demographic attributes into a structured representation, enabling personalized ECG synthesis.
    \item We propose \textbf{MIDT-ECG, a training paradigm} that augments standard MSE denoising with multi-resolution Mel-spectrogram supervision, enforcing clinically relevant morphology.
    \item We constructed a \textbf{comprehensive evaluation benchmark} across multiple dimensions, including synthetic signal fidelity, trustworthiness, inter-lead correlation, outlier analysis, data augmentation, and substitution scenarios for downstream tasks.
\end{itemize}
These contributions collectively establish a robust methodology for synthesizing trustworthy, personalized medical time series, providing a scalable and privacy-preserving foundation for cardiovascular AI research.
\section{Related Work}

\subsection{The Evolution of Generative Models for ECG Synthesis}
The synthesis of ECG signals has progressed through several generations of deep learning models. Early pioneering work utilized Generative Adversarial Networks (GANs) to produce single-lead waveforms~\citep{delaney2019synthesis}, with subsequent architectural improvements incorporating LSTMs and attention to better capture beat-by-beat morphology~\citep{zhu2019electrocardiogram, rafi2022heartnet}. Variational Autoencoders (VAEs) were also explored for their ability to learn structured latent spaces for data augmentation~\citep{sang2022generation, kuznetsov2021interpretable}. While foundational, these methods often struggled with training instability and capturing the long-range temporal dependencies of cardiac signals. More recently, diffusion models~\citep{ho2020denoising} have become the state-of-the-art, offering superior sample quality and stability. Architectures like SSSD-ECG~\citep{alcaraz2023diffusion}, which leverages Structured State-Space Models, and DiffuSETS~\citep{lai2025diffusets}, which uses a flexible multimodal conditioning mechanism, have demonstrated impressive results. However, despite this rapid architectural progress, two fundamental limitations persist: a reliance on simplistic time-domain training objectives and coarse-grained, unimodal conditioning, which leave the critical gaps in morphological fidelity and personalization unaddressed.

\subsection{Ensuring Trustworthiness in Generative Health AI}
The increasing deployment of generative AI in healthcare has created an urgent need for robust methods to ensure trustworthiness and manage risk. This is a broad challenge, with parallel efforts in synthesizing other private medical data, such as electroencephalography (EEG) signals for neurological applications~\citep{zhou2023generative, liu2024eeg2video} and complex, longitudinal electronic health records (EHR)~\citep{choi2017generating, he2023meddiff}. A key theme emerging from this work is the need for rigorous, standardized evaluation. Systematic benchmarks are being developed to assess the privacy and utility of synthetic tabular data~\citep{kaabachi2025scoping}, but a similar comprehensive framework for high-dimensional, clinically complex time-series data like the 12-lead ECG remains an open challenge. Our work contributes to this area by proposing and executing a multi-faceted evaluation protocol that explicitly measures signal fidelity, physiological coherence, privacy risk, and downstream clinical utility.

\subsection{Bridging the Fidelity Gap: From Time-Domain to Frequency-Domain}
The predominant training paradigm for time-series generation relies on time-domain losses like Mean Squared Error, which are often insufficient to enforce the morphological coherence essential for clinical realism. The field of audio synthesis, which faces similar challenges in capturing perceptual quality, has long demonstrated the power of frequency-domain supervision~\citep{sheng2019reducing,vasquez2019melnet}. This principle is beginning to be explored for ECGs, with concurrent work like ECG-DPM~\citep{10925073} using spectrogram-based diffusion models, but is based on UNet backbone and is not conditional. Our work introduces \textbf{MIDT-ECG}, a framework that applies a mel-spectrogram informed training paradigm, showing its suitability for ECGs through emphasis on low frequency bands, and provides the first rigorous evaluation of its impact on physiological coherence (e.g., interlead correlations) and its role as a surrogate for real data in data scarce settings. This bridges a methodological gap by imposing a stronger, clinically relevant structural prior on the generated waveforms.

\section{Methods}

\begin{figure}[ht]
    \centering
    % Page=1 selects the first page of the PDF
    \includegraphics[width=\textwidth,page=1]{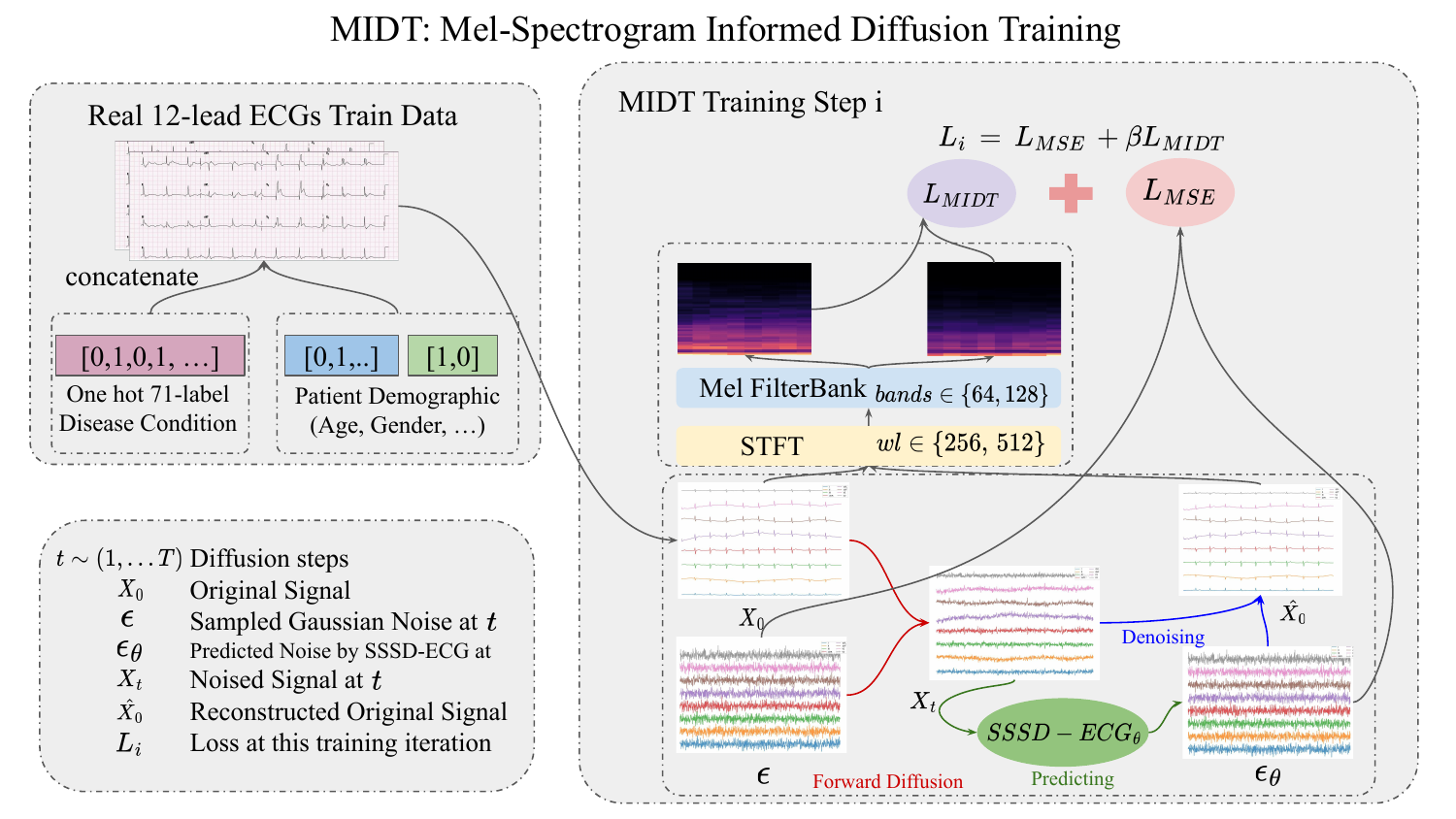}
    \caption{Mel-Spectrogram Informed Diffusion Training Overview. This diagram illustrate how MIDT works in on single MIDT training Step. Signal illustrations of $X_0, \epsilon, X_t, \epsilon_\theta, \hat{X_0} $ on the right grey box are generated by our trained model.}
    \label{fig:midtoverview}
\end{figure}

Our methodology illustrated in \ref{fig:midtoverview} enhances a state-of-the-art generative model for ECGs, addressing key limitations in morphological fidelity and personalization. We build upon the Structured State Space Diffusion (SSSD-ECG) model, introducing two targeted modifications: a novel training framework to improve waveform realism and an enhanced conditioning mechanism to enable patient-specific synthesis.

% \subsection{Foundation: The SSSD-ECG Model}

We select SSSD-ECG~\citep{alcaraz2023diffusion} as our foundational architecture due to its proven success in generating high-fidelity 12-lead ECGs. The model leverages a score-based diffusion process to iteratively transform random noise into structured signals. Its core strength lies in its use of Structured State-Space Model (SSSM) layers, which are highly effective at capturing the long-range temporal dependencies crucial for modeling the physiological structure of an entire heartbeat and rhythm. In its original implementation, SSSD-ECG conditioned on a 71 length onehot vector representing diagnostic labels, which is projected into a continuous representation via a learnable weight matrix.

Despite its strong performance, the original SSSD-ECG framework has two primary limitations. First, its reliance on a mean squared error (MSE) loss treats each time step independently, failing to impose a global structural prior on the waveform's morphology. Second, its conditioning is limited to a single-vector representation of disease labels, which prevents the generation of personalized ECGs that reflect individual patient attributes like age or gender.

% \subsection{Improving Morphological Fidelity with a Mel-Spectrogram Loss}

% To overcome the insufficiency of point-wise losses like MSE, we enforce structural consistency in the time-frequency domain. An ECG's diagnostic information is encoded in the shape and timing of its components, which have distinct spectral characteristics (e.g., the high-frequency QRS complex vs. the lower-frequency T-wave). A purely time-domain loss struggles to balance these components correctly. We address this by introducing an auxiliary loss based on the multi-resolution mel-spectrogram of the signal.

% This is not merely an additional loss term, but a mechanism to impose a strong prior on the signal's temporal coherence and morphological structure. We compute Short-Time Fourier Transforms (STFT) with multiple window sizes to capture both rapid, transient events (with small windows) and slower, evolving waveforms (with large windows). The resulting spectrograms are converted to the mel scale to focus on perceptually relevant frequency bands. The total training objective becomes a weighted sum of the time-domain and frequency-domain losses: $\mathcal{L}_{\text{Total}} = \mathcal{L}_{\text{MSE}} + \beta \mathcal{L}_{\text{Mel}}(\hat{y}, y)$, where $\mathcal{L}_{\text{Mel}}$ is the $L_1$ distance between the multi-resolution spectrograms of the generated ($\hat{y}$) and ground-truth ($y$) signals. This principled, frequency-domain supervision forces the model to learn not just individual data points, but the holistic structure of a clinically plausible ECG.
\subsection{MIDT-ECG: Mel-Spectrogram Informed Diffusion Training for Morphological Fidelity}

% \begin{algorithm}[t]
% \caption{Training with Mel-Spectrogram Supervision}\label{algorithm}
% \begin{algorithmic}[1]
% \Repeat
%     \State $x_{0} \sim q(x_{0})$ \Comment{sample real ECG}
%     \State $t \sim \text{Uniform}(\{1,\dots,T\})$ \Comment{sample diffusion step}
%     \State $\epsilon \sim \mathcal{N}(0,I)$ \Comment{sample Gaussian noise}
%     \State $x_{t} \gets \sqrt{\bar{\alpha}_{t}}\,x_{0} + \sqrt{1-\bar{\alpha}_{t}}\,\epsilon$ \Comment{forward diffusion}
%     \State $\epsilon_{\theta} \gets \text{net}(x_{t}, t, \text{cond\_vector})$ \Comment{predict noise}
%     \State $\hat{x}_{0} \gets \dfrac{x_{t} - \tfrac{1-\alpha_{t}}{\sqrt{1-\bar{\alpha}_{t}}}\,\epsilon_{\theta}}{\sqrt{\alpha_{t}}}$ \Comment{reconstruct clean signal}
%     \State $\mathcal{L}_{\text{MSE}} \gets \|\epsilon - \epsilon_{\theta}\|^{2}$ \Comment{denoising loss}
%     \State $\mathcal{L}_{\text{MIDT}} \gets \| \text{Mel}(\hat{x}_{0}) - \text{Mel}(x_{0}) \|_{1}$ \Comment{multi-resolution Mel-spectrograms}
%     \State $\mathcal{L}_{\text{Total}} \gets \mathcal{L}_{\text{MSE}} + \beta \,\mathcal{L}_{\text{MIDT}}$
%     \State Take gradient descent step on $\nabla_{\theta} \,\mathcal{L}_{\text{Total}}$
% \Until{converged}
% \end{algorithmic}
% \end{algorithm}

A primary limitation of standard diffusion training is its reliance on point-wise losses like MSE, which are agnostic to the underlying temporal structure of a signal. For an ECG, where diagnostic information is encoded in the shape and duration of waveform components, an MSE loss is insufficient. To overcome this, we introduce MIDT-ECG (Mel-Spectrogram Informed Diffusion Training for ECGs), a principled paradigm that supervises the model in the time-frequency domain.

by Mel Spectrogram used as a higher fidelity of continuous representation other than vector quantization in audio synthesis \citep{meng2024autoregressive} but is specifically adapted to the unique physiological characteristics of the ECG.  To capture both high-frequency transients (e.g., QRS complexes) and low-frequency dynamics (e.g., T-waves), we compute multi-resolution Short-Time Fourier Transforms (STFTs) using multiple window sizes. Beyond the standard MSE denoising loss, we incorporate a Mel-spectrogram loss $\mathcal{L}{\text{MIDT}}$ that aligns multi-resolution Mel-spectrograms of real and reconstructed ECGs, encouraging the network to preserve waveform morphology and ensuring both pointwise fidelity and physiological coherence.

Crucially, we then warp the frequency axis of these spectrograms onto the perceptually-motivated Mel scale. The Mel scale's non-linear compression of frequencies is uniquely suited for ECG analysis, as it naturally places greater emphasis on the diagnostically-rich low-frequency bands where information about ST segments and T-wave morphology resides. This imposes a strong inductive bias, forcing the model to prioritize the most clinically relevant spectral components. We calculate $\mathcal{L}_{\text{MIDT}}$ as the $L_1$ distance between the multi-resolution mel-spectrograms~\cite{kumar2023high}. The final training objective is a weighted sum: 
\begin{equation}
    \mathcal{L}_{\text{Total}} = \mathcal{L}_{\text{MSE}} + \beta \mathcal{L}_{\text{MIDT}}(\hat{y}, y) 
\end{equation}
where $\mathcal{L}_{\text{MSE}}$ is the pointwise mean sqaure error (MSE) between signals. $\mathcal{L}_{\text{MIDT}}$ is not merely an auxiliary loss, but a core training mechanism that forces the model to learn the holistic, clinically plausible structure of an ECG. 

\subsection{Enabling Personalization with Disentangled Multimodal Conditioning}
To enable patient-specific synthesis, we developed an enhanced conditioning mechanism designed to learn a disentangled mapping from a patient's profile to their electrophysiological signature.
The key idea is to move away from representing all patient information with a single conditioning vector, which compresses heterogeneous attributes into a single undifferentiated embedding. Instead, we build a structured and disentangled representation of patient attributes.

Concretely, we partition the conditioning inputs into distinct groups that capture complementary sources of variability, such as diagnostic categories, rhythm labels, and demographic information (e.g., age bins, gender). Each attribute group is first encoded as a one-hot vector $\mathbf{y}_k$, which is then projected into its own continuous embedding space $\mathbf{e}_k$ using a dedicated weight matrix $\mathbf{W}_k:\mathbf{W}_k^\top \mathbf{y}_k$.

These disentangled embeddings are subsequently concatenated into a single, comprehensive patient representation vector, $\mathbf{c} = \text{Concat}(\mathbf{e}_{\text{diag}}, \dots, \mathbf{e}_{\text{age}}, \dots)$. This patient representation $\mathbf{c}$ serves as a patient-specific prior that conditions the entire reverse diffusion process. Rather than being injected only once, $\mathbf{c}$ is provided to every layer of the SSSD network, where it modulates internal activations. This layer-wise conditioning ensures that the generated ECG signals are consistent not only with general disease classes but also with the finer-grained physiological nuances of the target patient profile, thereby enabling personalized synthesis.

Specifically, for multimodal conditioning, input features were organized into clinically meaningful categories: 
\begin{itemize}[leftmargin=*]
    \item \textbf{Clinical Labels:} The 71 SCP statement labels were grouped into three categories: Diagnostic (40 labels, e.g., \texttt{MI}), Form (19 labels, e.g., \texttt{HVOLT}), and Rhythm (12 labels, e.g., \texttt{AFIB}). Each group was one-hot encoded independently.  
    \item \textbf{Demographic Features:} Continuous demographic variables were discretized into clinically relevant bins and then one-hot encoded:  
    \begin{itemize}
        \item \textbf{Age:} 6 bins defined by cutoffs at [12, 17, 34, 54, 74].  
        \item \textbf{Gender:} 2 classes (male, female).  
        % \item \textbf{BMI:} 6 bins based on standard clinical cutoffs [18.5, 25, 30, 35, 40].  
    \end{itemize}
\end{itemize}
All one-hot vectors were subsequently projected into a shared 32-dimensional embedding space, forming the final conditioning representation.
% This vector serves as a powerful patient-specific prior that guides the entire reverse diffusion process. It is injected into each SSSD layer, where it modulates the network's internal activations, ensuring the generated signal is consistent not only with a general disease class but with the specific physiological nuances of the target patient profile.

\section{Experiments and Results}
To rigorously validate our proposed methods, we designed a multi-faceted evaluation framework on the public PTB-XL dataset~\citep{wagner2020ptb}. This public dataset contains 21,837 clinical 12-lead ECG recordings from 18,885 patients. Each 10-second recording was sampled at 100 Hz (1,000 time steps per lead). We used the standard patient-level data splits to ensure no data leakage, resulting in 17,441 training, 2,193 validation, and 2,203 test samples. Patient demographic information (age, gender, height, weight) was extracted from the metadata to enable personalized conditioning.

Our investigation is structured around a central question for deploying synthetic data in healthcare: Are the generated signals trustworthy, useful, and robust? To answer these, we conduct a systematic comparative analysis within a controlled experimental tested based on the SSSD-ECG architecture. This analysis includes: (i) the unmodified baseline model SSSG-ECG; (ii) our proposed \textbf{MIDT-ECG} framework; and (iii) demographic-conditioned variants which combine our multimodal conditioning, such as \textbf{SSSD-ECG+A} for age and \textbf{SSSD-ECG+G} for gender. This design allows us to systematically quantify the impact of our enhancements.

To capture overall signal quality, we include statistical fidelity metrics such as RMSE, MSE, and Signal-to-Noise Ratio (SNR). To go beyond point-wise similarity, we also assess morphological realism using Fourier distance, Hausdorff distance, and SSIM, which quantify global waveform structure and shape consistency. Finally, because privacy preservation is critical in synthetic healthcare data, we report two complementary privacy metrics: Membership Inference Risk (MIR) and Nearest-Neighbor Adversarial Accuracy (NNAA). Together, these metrics allow us to evaluate not only the fidelity of the generated signals but also their clinical realism and privacy robustness.

% \begin{table}[h!]
% \centering
% \caption{Comprehensive comparison of signal fidelity, morphology, and privacy. The MIDT-ECG framework excels in morphological realism and offers the strongest privacy guarantees.}
% \label{tab:fidelity_and_privacy}
% \vspace{0.5em}
% \resizebox{\textwidth}{!}{%
% \begin{tabular}{@{}lcccccccc@{}}
% \toprule
% & \multicolumn{6}{c}{\textbf{Fidelity \& Morphology Metrics}} & \multicolumn{2}{c}{\textbf{Privacy Metrics}} \\
% \cmidrule(lr){2-7} \cmidrule(lr){8-9}
% \textbf{Training Objective} & \textbf{RMSE} $\downarrow$ & \textbf{MSE} $\downarrow$ & \textbf{SNR (dB)} $\uparrow$ & \textbf{Fourier} $\downarrow$ & \textbf{Hausdorff} $\downarrow$ & \textbf{SSIM} $\uparrow$ & \textbf{MIR} $\downarrow$ & \textbf{NNAA} $\downarrow$ \\
% \midrule
% SSSD-ECG     & 0.2114 & 0.0524 & -3.086 & 0.2115 & 1.1870 & 0.6004 & 0.0099 & 0.0047 \\
% SSSD-ECG+A     & 0.2145 & 0.0641 & \textbf{-1.288} & 0.2145 & 1.1889 & 0.6090 & 0.0045 & 0.0178 \\
% SSSD-ECG+G &0.2846&0.0975&-4.594&0.2845&1.5284&0.6100&\textbf{0.0036}&0.0026 \\
% MIDT-ECG  & \textbf{0.2015} & \textbf{0.0501} & -2.508 & \textbf{0.2016} & \textbf{1.0860} & \textbf{0.6313} & 0.0081 & \textbf{-0.0009} \\
% % D+Multi     & 0.2593 & 0.0852 & -3.265 & 0.2593 & 1.3242 & 0.5979 & -- & -- \\
% % D+BMI     & 0.2408 & 0.0768 & -2.443 & 0.2408 & 1.2830 & 0.5795 & -- & -- \\
% \bottomrule
% \end{tabular}
% }
% \end{table}
% in preamble:
% \usepackage{booktabs,makecell,xcolor}

% Preamble
% \usepackage{booktabs,xcolor}
\newcommand{\goodd}[1]{\textcolor{teal!70!black}{#1}}
\newcommand{\badd}[1]{\textcolor{red!70!black}{#1}}
\newcommand{\deltag}[1]{\textsuperscript{\scriptsize\;\goodd{(Δ #1)}}}
\newcommand{\deltab}[1]{\textsuperscript{\scriptsize\;\badd{(Δ #1)}}}
% Usage: choose deltag (green) when better than baseline, deltab (red) when worse.

\begin{table}[t]
\centering
\caption{Comparative evaluation of signal fidelity, morphological quality, and privacy. Values are shown with deltas ($\Delta$) relative to the baseline (SSSD-ECG); improvements highlighted in green, degradations in red.}
\label{tab:fidelity_and_privacy}
\vspace{0.25em}
\renewcommand{\arraystretch}{1.15}
\resizebox{\textwidth}{!}{%
\begin{tabular}{@{}lcccccccc@{}}
\toprule
& \multicolumn{6}{c}{\textbf{Fidelity \& Morphology Metrics}} & \multicolumn{2}{c}{\textbf{Privacy Metrics}} \\
\cmidrule(lr){2-7} \cmidrule(lr){8-9}
\textbf{Training Objective} & \textbf{RMSE} $\downarrow$ & \textbf{MSE} $\downarrow$ & \textbf{SNR (dB)} $\uparrow$ & \textbf{Fourier} $\downarrow$ & \textbf{Hausdorff} $\downarrow$ & \textbf{SSIM} $\uparrow$ & \textbf{MIR} $\downarrow$ & \textbf{NNAA} $\downarrow$ \\
\midrule
\textbf{SSSD-ECG (Baseline)} & 0.2114 & 0.0524 & -3.086 & 0.2115 & 1.1870 & 0.6004 & 0.0099 & 0.0047 \\
\midrule
SSSD-ECG + A
& 0.2145 \deltab{+0.0031}
& 0.0641 \deltab{+0.0117}
& \textbf{-1.288} \deltag{+1.798}
& 0.2145 \deltab{+0.0030}
& 1.1889 \deltab{+0.0019}
& 0.6090 \deltag{+0.0086}
& 0.0045 \deltag{-0.0054}
& 0.0178 \deltab{+0.0131}
\\
SSSD-ECG + G
& 0.2846 \deltab{+0.0732}
& 0.0975 \deltab{+0.0451}
& -4.594 \deltab{-1.508}
& 0.2845 \deltab{+0.0730}
& 1.5284 \deltab{+0.3414}
& 0.6100 \deltag{+0.0096}
& \textbf{0.0036} \deltag{-0.0063}
& 0.0026 \deltag{-0.0021}
\\
MIDT-ECG (Ours)
& \textbf{0.2015} \deltag{-0.0099}
& \textbf{0.0501} \deltag{-0.0023}
& -2.508 \deltag{+0.578}
& \textbf{0.2016} \deltag{-0.0099}
& \textbf{1.0860} \deltag{-0.1010}
& \textbf{0.6313} \deltag{+0.0309}
& 0.0081 \deltag{-0.0018}
& \textbf{-0.0009} \deltag{-0.0056}
\\
% MIDT-ECG + A (Ours)
% & \textbf{0.2015} \deltag{-0.0099}
% & \textbf{0.0501} \deltag{-0.0023}
% & -2.508 \deltag{+0.578}
% & \textbf{0.2016} \deltag{-0.0099}
% & \textbf{1.0860} \deltag{-0.1010}
% & \textbf{0.6313} \deltag{+0.0309}
% & 0.0081 \deltag{-0.0018}
% & \textbf{-0.0009} \deltag{-0.0056}
% \\
% MIDT-ECG + G (Ours)
% & \textbf{0.2015} \deltag{-0.0099}
% & \textbf{0.0501} \deltag{-0.0023}
% & -2.508 \deltag{+0.578}
% & \textbf{0.2016} \deltag{-0.0099}
% & \textbf{1.0860} \deltag{-0.1010}
% & \textbf{0.6313} \deltag{+0.0309}
% & 0.0081 \deltag{-0.0018}
% & \textbf{-0.0009} \deltag{-0.0056}
% \\
\bottomrule
\end{tabular}
}
\end{table}

\subsection{Signal Fidelity and Trustworthiness.}
We first established the foundational quality and trustworthiness of the signals. A comprehensive comparison of signal fidelity, morphological realism, and privacy preservation is provided in Table~\ref{tab:fidelity_and_privacy}. The results reveal a clear separation of benefits. Applying multimodal conditioning with age (SSSD-ECG+A) significantly improves the Signal-to-Noise Ratio (SNR), indicating better physiological amplitude scaling. However, the greatest improvement in morphological accuracy comes from our proposed \textbf{MIDT-ECG} framework across various fidelity and morphology metrics. For example, it reduces RMSE by 5\% and Hausdorff distance by nearly 9\% relative to the baseline. Crucially, these improvements are not achieved at the expense of privacy: MIDT-ECG attains the lowest MIR and NNAA scores, demonstrating reduced risk of membership inference and nearest-neighbor leakage. These facts position it as the most reliable framework for patient-specific ECG synthesis.

% \sisetup{
%   table-format=1.3,
%   table-number-alignment=center,
%   detect-weight=true,
%   detect-inline-weight=math
% }

% \begin{wraptable}{r}{0.4\textwidth}
%     \centering
%     \vspace{-0.75em} % tighter to preceding text
%     \caption{Average and maximum absolute inter-lead correlation error relative to real data.}
%     \label{tab:corr_error_summary}
%     \vspace{0.25em}
%     \resizebox{\linewidth}{!}{%
%         \begin{tabular}{@{}lSS@{}}
%         \toprule
%         \textbf{Model} & {\textbf{Avg. Corr.}$\downarrow$} & {\textbf{Max Corr.}$\downarrow$} \\
%         \midrule
%         SSSD-ECG (Baseline) & 0.140 & 0.491 \\
%         SSSD-ECG+A          & 0.124 & 0.456 \\
%         SSSD-ECG+G          & 0.267 & 0.899 \\
%         MIDT-ECG (Ours)     & \textbf{0.042} & \textbf{0.108} \\
%         \bottomrule
%         \end{tabular}
%     }
%     \vspace{-0.5em} % reduce gap below table
% \end{wraptable}

% Preamble:
% \usepackage{booktabs,siunitx,xcolor,caption,wrapfig}
\sisetup{
  table-format=1.3,
  table-number-alignment=center,
  detect-weight=true,
  detect-inline-weight=math
}
\renewcommand{\goodd}[1]{\textcolor{teal!70!black}{#1}}
\renewcommand{\badd}[1]{\textcolor{red!70!black}{#1}}

% A tiny right-aligned "delta" text column (no extra padding)
\newcolumntype{d}{@{}>{\tiny}r@{}}

\begin{wraptable}{r}{0.5\textwidth}
  \centering
  \vspace{-0.7em}
  \caption{Average and maximum absolute inter-lead correlation error vs. real data. Δ is relative to SSSD-ECG (Baseline).}
  \label{tab:corr_error_summary}
  \vspace{0.25em}
  \resizebox{\linewidth}{!}{%
    \begin{tabular}{@{}l
    S[round-mode=places, round-precision=3]
    d
    S[round-mode=places, round-precision=3]
    d @{}}
      \toprule
      \textbf{Model}
      & \multicolumn{2}{c}{\textbf{Avg. Corr.}\,$\downarrow$}
      & \multicolumn{2}{c}{\textbf{Max Corr.}\,$\downarrow$} \\
      \cmidrule(lr){2-3}\cmidrule(lr){4-5}
      %         value     delta            value     delta
      SSSD-ECG (Baseline) & 0.140 &        & 0.491 &        \\
      \midrule
      SSSD-ECG + A          & 0.124 & \deltag{-0.016}\, & 0.456 & \deltag{-0.035} \\
      SSSD-ECG + G          & 0.267 & \deltab{+0.127}\, & 0.899 & \deltab{+0.408} \\
      MIDT-ECG (Ours)     & \textbf{0.042} & \deltag{-0.098}\, & \textbf{0.108} & \deltag{-0.383} \\
      \bottomrule
    \end{tabular}
  }
  \vspace{-0.5em}
\end{wraptable}

To further assess physiological coherence, we analyzed the inter-lead correlations, a critical property of realistic ECGs. The results are shown in Table~\ref{tab:corr_error_summary}. The \textbf{MIDT-ECG} framework demonstrates a 70\% reduction in the average absolute correlation error, from 0.140 down to 0.042. This confirms its superior ability to capture the complex spatio-temporal dependencies between leads, a key aspect of clinical realism. This establishes that our proposed method produces signals that are not only more accurate but also more physiologically plausible and trustworthy.

\subsection{Inter-lead Correlation Analysis}
\label{app:correlation}
A fundamental property of clinically valid 12-lead ECGs is the complex set of physiological correlations between different leads, which reflect the three-dimensional propagation of the heart's electrical wavefront. A high-fidelity generative model must successfully capture these spatio-temporal relationships. To visually and quantitatively assess this, we computed Pearson correlation matrices for real and synthetic data and visualized them as heatmaps. The following figures provide a detailed comparison.

Figure~\ref{fig:heatmap_real} shows the ground-truth correlation matrix computed from real ECGs in the PTB-XL test set. It displays well-known clinical patterns, such as the strong positive correlation between adjacent precordial leads (e.g., V1-V2) and the characteristic negative correlation between limb leads I and III. This serves as the reference against which the synthetic models are compared.

Figures~\ref{fig:heatmap_synth_raw} and \ref{fig:heatmap_synth_mel} show the correlation matrices for the synthetic data generated by the baseline SSSD-ECG model and our proposed MIDT-ECG framework, respectively. A visual inspection reveals that while the SSSD-ECG model captures the general structure, the MIDT-ECG's matrix is a much closer match to the ground truth in Figure~\ref{fig:heatmap_real}.

\begin{figure}[h!]
    \centering
    \begin{subfigure}[t]{0.31\textwidth}
        \centering
        \includegraphics[width=\linewidth]{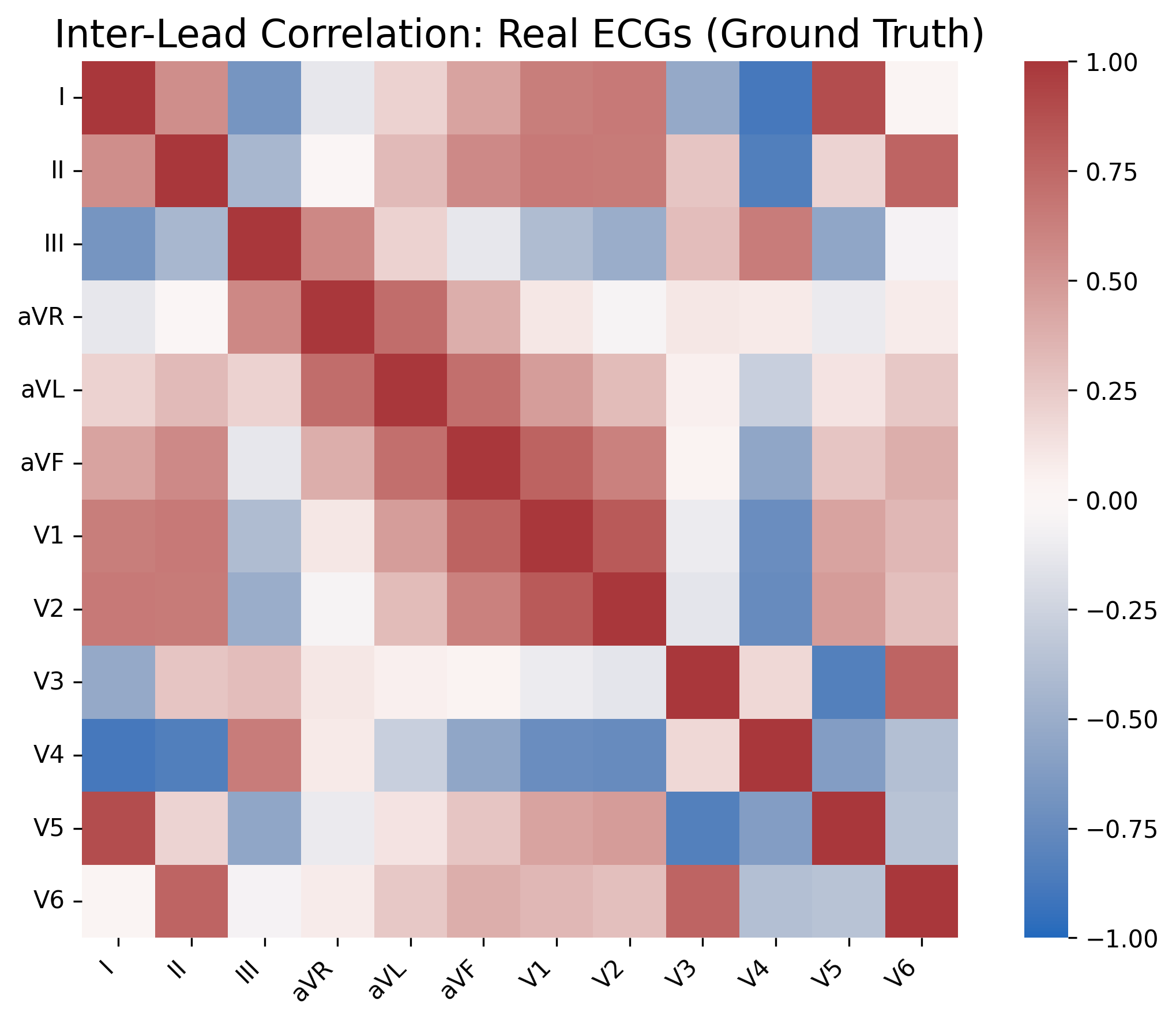}
        \caption{The ground-truth inter-lead correlation matrix for real ECG data.}
        \label{fig:heatmap_real}
    \end{subfigure}
    \hfill 
    \begin{subfigure}[t]{0.31\textwidth}
        \centering
        \includegraphics[width=\linewidth]{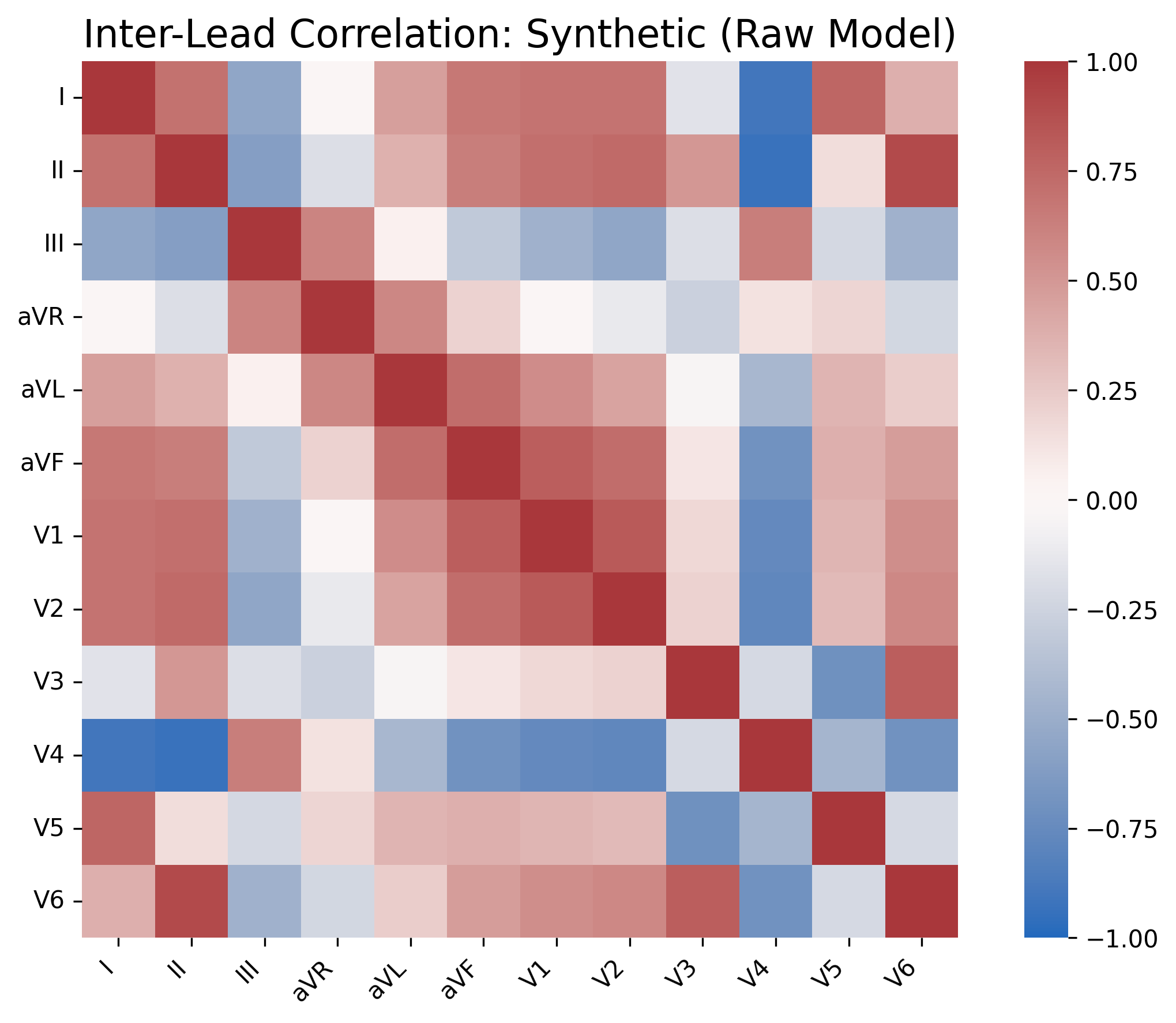}
        \caption{Inter-lead correlation matrix for synthetic data from the baseline \texttt{SSSD-ECG} model.}
        \label{fig:heatmap_synth_raw}
    \end{subfigure}
    \hfill 
    \begin{subfigure}[t]{0.32\textwidth}
        \centering
        \includegraphics[width=\linewidth]{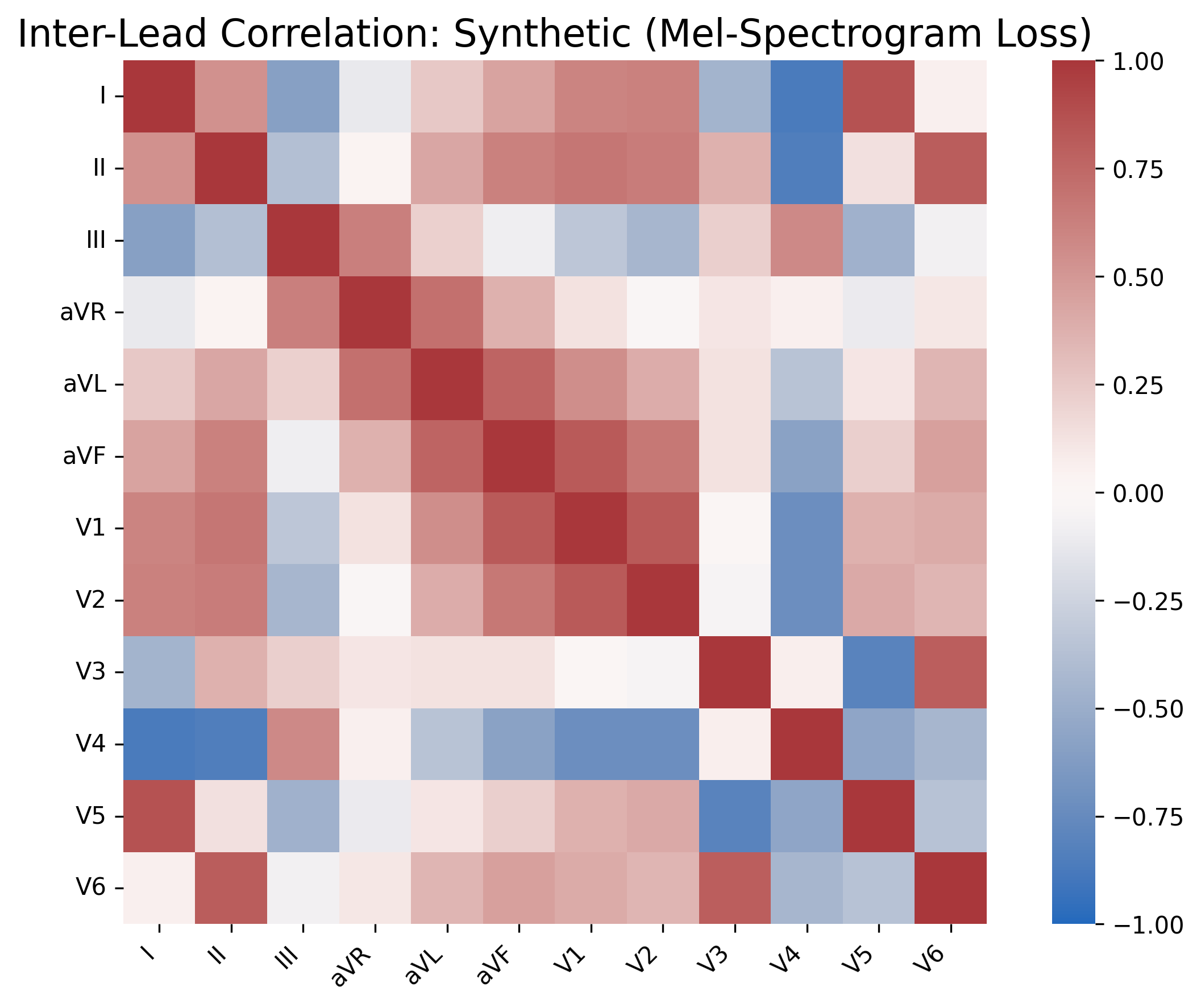}
        \caption{Inter-lead correlation matrix for synthetic data from our MIDT-ECG framework.}
        \label{fig:heatmap_synth_mel}
    \end{subfigure}
    \caption{Comparison of inter-lead correlation matrices for real and synthetic ECG data.}
    \label{fig:heatmaps_combined}
\end{figure}

\begin{figure}[h!]
    \centering
    % Change [b] or the default to [t] for TOP alignment
    \begin{subfigure}[t]{0.23\textwidth}
        \centering
        \includegraphics[width=\linewidth]{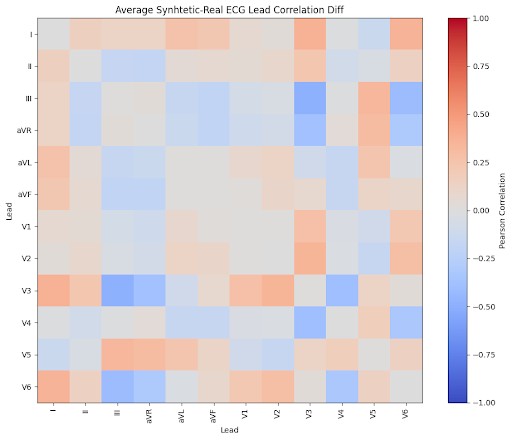}
        \caption{Difference matrix (Real - SSSD-ECG Baseline).}
        \label{fig:heatmap_diff_raw}
    \end{subfigure}
    \hfill 
    % Also use [t] here
    \begin{subfigure}[t]{0.23\textwidth}
        \centering
        \includegraphics[width=\linewidth]{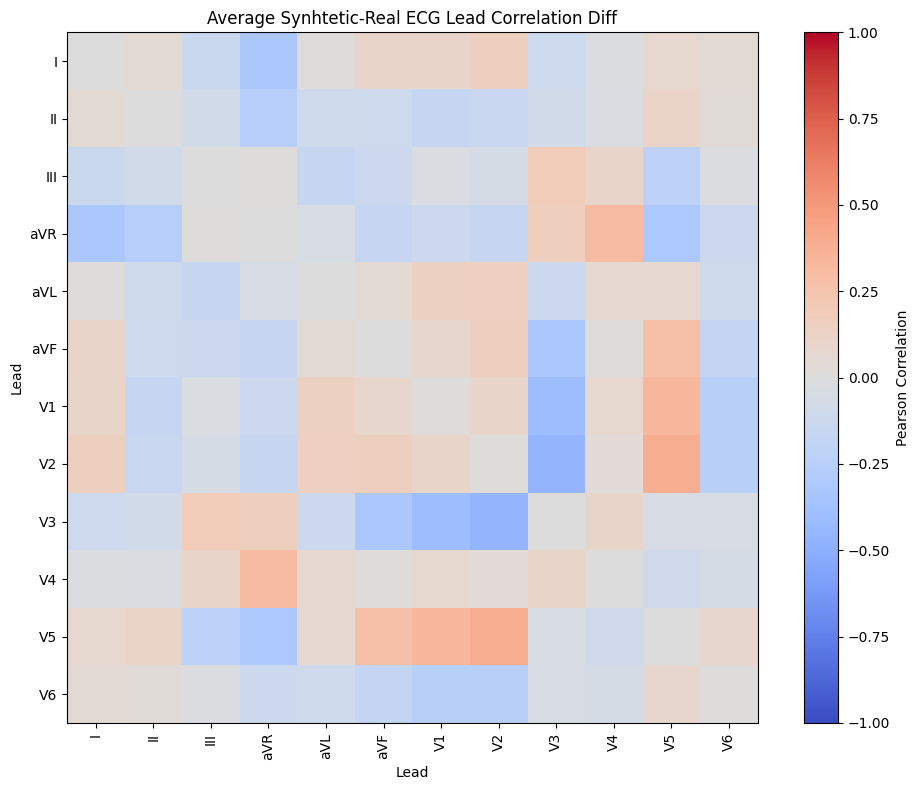}
        \caption{Difference matrix (Real - SSSD-ECG+A).}
        \label{fig:heatmap_diff_mel}
    \end{subfigure}
    \hfill
    \begin{subfigure}[t]{0.23\textwidth}
        \centering
        \includegraphics[width=\linewidth]{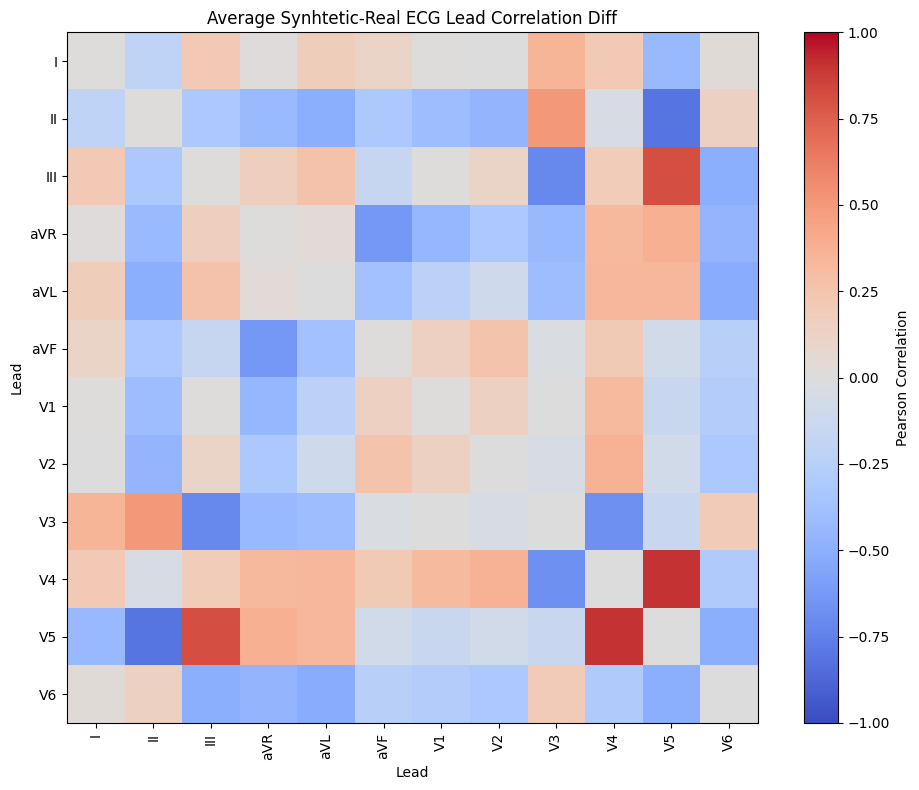}
        \caption{Difference matrix (Real - SSSD-ECG+G).}
        \label{fig:heatmap_diff_mel}
    \end{subfigure}
    \hfill
    \begin{subfigure}[t]
    {0.23\textwidth}
        \centering
        \includegraphics[width=\linewidth]{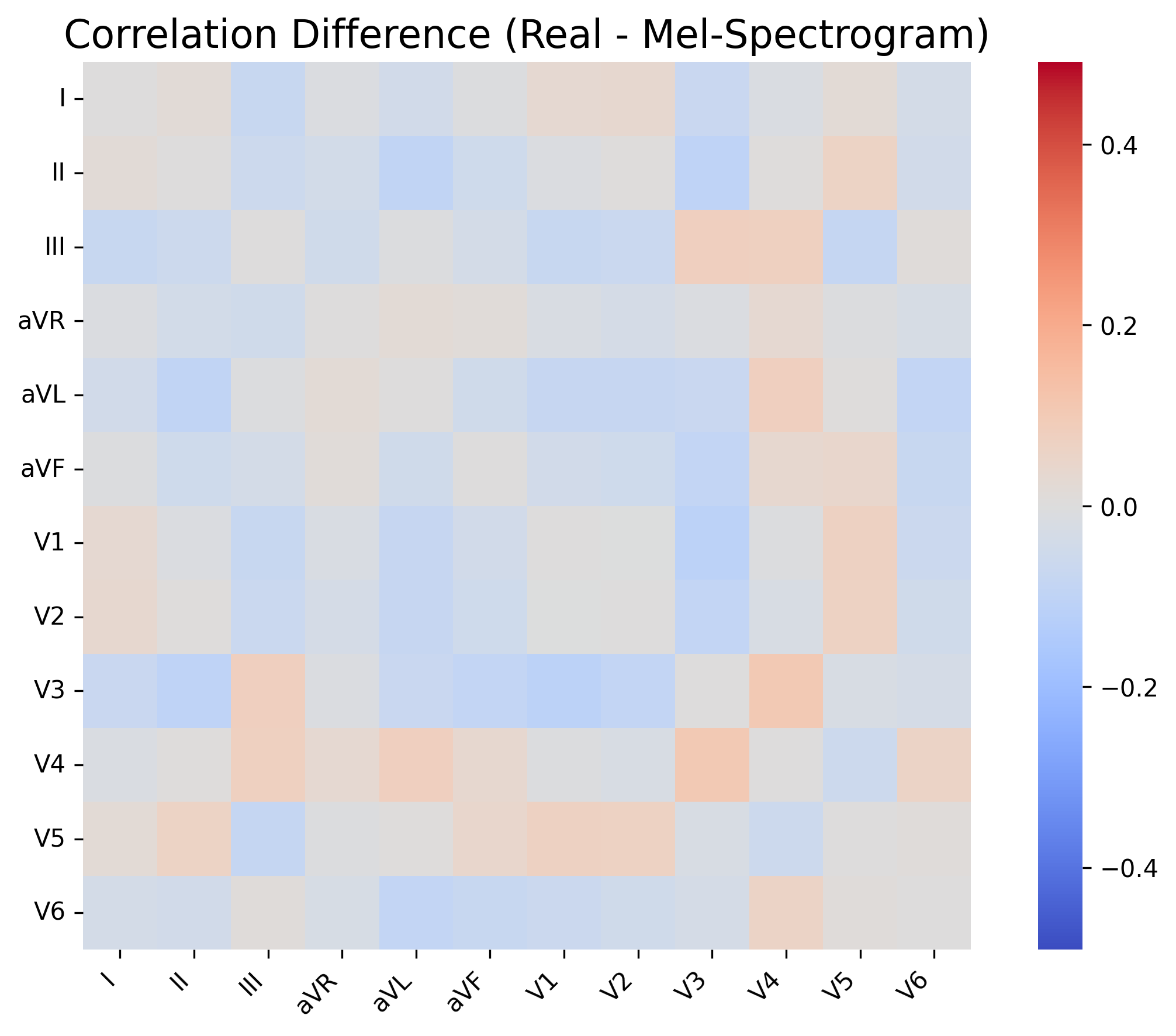}
        \caption{Difference matrix (Real - MIDT-ECG (Ours).}
        \label{fig:heatmap_diff_mel}
    \end{subfigure}
    \caption{Comparison of correlation difference matrices for the baseline and proposed models. Darker colors indicate larger errors, 1 is positive relevant, -1 is negative relevant. The significantly paler colors demonstrate the superior performance of our MIDT-ECG method.}
    \label{fig:diff_heatmaps_combined}
\end{figure}

Furthermore, the superiority of the MIDT-ECG framework is confirmed by the difference heatmaps in Figures~\ref{fig:heatmap_diff_raw} and \ref{fig:heatmap_diff_mel} as well. The difference matrix for the SSSD-ECG model (Figure~\ref{fig:heatmap_diff_raw}) shows large error patches (darker reds and blues), indicating a significant deviation from the real data's physiological structure. In stark contrast, the difference matrix for the MIDT-ECG framework (Figure~\ref{fig:heatmap_diff_mel}) is substantially more muted and closer to the neutral zero-centered color, indicating a much smaller error. This visual evidence provides a clear intuition for the quantitative results reported in the main paper, where the MIDT-ECG framework reduced the average absolute correlation error by 70\%. This analysis provides compelling evidence that the frequency-domain supervision of the Mel-spectrogram loss is crucial for generating ECGs that are not only morphologically accurate but also physiologically coherent.

\subsection{Data Augmentation Scenarios}
\label{app:augmentation_analysis}

To fully understand the utility and robustness of our generative models, we conducted two complementary data augmentation experiments. The first investigates the marginal value of synthetic data in a data-rich environment, while the second (presented in the main paper) tests its role as a surrogate in data-scarce environments. This appendix provides the full results and a detailed analysis of both scenarios.

\subsubsection{Augmenting a Complete Real Dataset (Data-Rich Environment)}
This experiment is designed to answer the question: "If I already have a sufficient amount of real data, can adding synthetic data provide any further benefit?" It evaluates the marginal utility of synthetic data by starting with a complete real dataset (8 folds) and incrementally adding folds of synthetic data.

The full results are presented in Table~\ref{tab:augmentation_impact_full}. The key finding is that performance gains are marginal and plateau quickly, indicating a point of diminishing returns for augmentation when real data is abundant. This is an expected and important result, as it confirms that a large real dataset is difficult to improve upon. However, even in this challenging scenario, the MIDT-ECG framework consistently demonstrates superior performance. While other models may have a slight edge with minimal augmentation (e.g., \texttt{disease + gender} at 1 fold), the MIDT-ECG framework achieves the highest AUROC scores as more data is added, culminating in the best overall Average Rank (1.50). This demonstrates its robustness and its ability to generate the most diagnostically useful signals, even in a context where their marginal contribution is small.

\begin{table}[t]
\centering
\caption{Full results of augmenting a complete real dataset (8 folds) with an increasing number of synthetic folds, measured by AUROC (mean ± 95\% CI).}
\label{tab:augmentation_impact_full}
\vspace{0.5em}
\resizebox{\textwidth}{!}{%
\begin{tabular}{@{}lccccccccc@{}}
\toprule
& \multicolumn{8}{c}{\textbf{Number of Synthetic Folds Added}} \\
\cmidrule(lr){2-9}
\textbf{Generator Type} & \textbf{1} & \textbf{2} & \textbf{3} & \textbf{4} & \textbf{5} & \textbf{6} & \textbf{7} & \textbf{8} & \textbf{Avg Rank} \\
\midrule
\textit{Synthetic Models} & & & & & & & & & \\
\quad SSSD-ECG (Baseline) & 0.928 ± 0.002 & 0.930 ± 0.002 & 0.929 ± 0.004 & 0.928 ± 0.003 & 0.928 ± 0.004 & 0.928 ± 0.003 & 0.926 ± 0.004 & 0.928 ± 0.002 & 3.00 \\
\quad SSSD-ECG+A & 0.928 ± 0.002 & 0.927 ± 0.006 & 0.928 ± 0.003 & 0.927 ± 0.002 & 0.926 ± 0.003 & \textbf{0.928 ± 0.004} & 0.927 ± 0.003 & 0.928 ± 0.003 & 4.38 \\
\quad SSSD-ECG+G & \textbf{0.930 ± 0.001} & 0.928 ± 0.003 & 0.925 ± 0.005 & 0.927 ± 0.003 & 0.926 ± 0.004 & 0.926 ± 0.004 & 0.928 ± 0.003 & 0.927 ± 0.002 & 3.75 \\
\quad MIDT-ECG (Ours) & 0.928 ± 0.002 & \textbf{0.931 ± 0.003} & \textbf{0.930 ± 0.004} & \textbf{0.929 ± 0.003} & \textbf{0.929 ± 0.005} & 0.928 ± 0.003 & \textbf{0.929 ± 0.004} & \textbf{0.931 ± 0.004} & \textbf{1.50} \\
\bottomrule
\end{tabular}
}
\end{table}

\begin{table}[h!]
\centering
\caption{Full results for the data substitution experiment, measured by AUROC (mean ± 95\% CI). *P value < 0.05 vs. best model in that column.}
\label{tab:substitution_full}
\vspace{0.5em}
\resizebox{\textwidth}{!}{%
\begin{tabular}{@{}lcccccccccc@{}}
\toprule
& \multicolumn{9}{c}{\textbf{Number of Real Data Folds Added}} \\
\cmidrule(lr){2-10}
\textbf{Data Type} & \textbf{0} & \textbf{1} & \textbf{2} & \textbf{3} & \textbf{4} & \textbf{5} & \textbf{6} & \textbf{7} & \textbf{8} & \textbf{Avg Rank} \\
\midrule
\textit{Baseline} & & & & & & & & & & \\
\quad Real Data Only & — & 0.901 ± 0.009 & \textbf{0.912 ± 0.003} & 0.916 ± 0.003 & \textbf{0.922 ± 0.005} & \textbf{0.924 ± 0.003} & \textbf{0.927 ± 0.002} & 0.926 ± 0.003 & 0.927 ± 0.005 & 2.62 \\
\midrule
\textit{Synthetic Models} & & & & & & & & & & \\
\quad Synthetic (SSSD-ECG) & 0.541 ± 0.074 & 0.901 ± 0.007 & \textbf{0.914 ± 0.002} & 0.917 ± 0.004 & 0.920 ± 0.004 & 0.923 ± 0.005 & 0.926 ± 0.005 & \textbf{0.928 ± 0.003} & 0.927 ± 0.005 & 2.89 \\
\quad Synthetic (SSSD-ECG+A) & 0.552 ± 0.068 & 0.901 ± 0.004 & 0.906 ± 0.007* & 0.914 ± 0.004 & 0.918 ± 0.006 & 0.922 ± 0.003 & 0.924 ± 0.004 & 0.925 ± 0.003 & 0.927 ± 0.002 & 5.00 \\
\quad Synthetic (SSSD-ECG+G) & 0.507 ± 0.067* & 0.894 ± 0.002* & 0.905 ± 0.002* & 0.911 ± 0.003* & 0.920 ± 0.003 & 0.923 ± 0.003 & 0.924 ± 0.003 & 0.923 ± 0.001* & 0.927 ± 0.003 & 5.11 \\
\quad Synthetic (MIDT-ECG) & \textbf{0.640 ± 0.094} & \textbf{0.902 ± 0.004} & 0.911 ± 0.002* & \textbf{0.919 ± 0.004} & 0.920 ± 0.005 & 0.923 ± 0.004 & 0.925 ± 0.002 & 0.926 ± 0.004 & \textbf{0.928 ± 0.002} & \textbf{2.78} \\
\bottomrule
\end{tabular}
}
\end{table}

\subsubsection{Synthetic Data as a Surrogate for Real Data (Data-Scarce Environment)}
This is the most critical use-case for synthetic data, designed to answer the question: "Can synthetic data substitute for real data when real data is unavailable or scarce?" To simulate this scenario, we begin with a fully synthetic dataset (8 folds) and incrementally add folds of real data, mimicking a researcher’s access to an expanding real-world cohort. The complete results are reported in Table~\ref{tab:substitution_full}.

The full results are presented in Table~\ref{tab:substitution_full}. This experiment yields three findings. First, when trained exclusively on synthetic data, the MIDT-ECG framework achieves an AUROC of 0.640, substantially outperforming the SSSD-ECG baseline (0.541). This demonstrates that our proposed spectral loss is essential for producing synthetic signals with meaningful diagnostic value. Second, in the critical low-data regime (1-3 folds), hybrid datasets combining synthetic and real data consistently match or surpass the performance of real-only baselines. This provides strong evidence that high-quality synthetic data can effectively bridge gaps in data availability. Third, as the amount of real data increases (4-8 folds), the performance of all methods converges towards the same upper bound. This confirms that when sufficient real data is available, it remains the gold standard, while synthetic data shifts from serving as a surrogate to acting as a supplementary resource.

\section{Discussion}
In this work, we introduced a comprehensive and clinically grounded benchmark for evaluating synthetic ECG generation models, focusing on four pillars: fidelity, personalization, privacy preservation, and clinical utility. Our study extends the capabilities of diffusion-based models, particularly SSSD-ECG, by incorporating demographic-aware conditioning and a mel-spectrogram-based loss to enhance morphological realism and signal coherence. Together, these contributions form a principled framework for both model development and evaluation.

\paragraph{A Structured Benchmarking Framework.}  
We propose a unified evaluation protocol that integrates statistical error metrics, morphological similarity, inter-lead correlation, clinical feature distributions, label faithfulness, and privacy risk. This framework moves beyond traditional point-wise metrics like RMSE or MSE, offering a multi-dimensional and clinically meaningful assessment of synthetic ECG quality. Importantly, by including real-vs-real baselines and inter-lead correlation analysis, we contextualize synthetic performance relative to natural physiological variability—helping distinguish true signal degradation from acceptable variability.

\paragraph{Faithfulness and Clinical Alignment.}  
A key innovation in our benchmark is the introduction of a faithfulness metric, which quantifies whether synthetic ECGs preserve label consistency when evaluated by classifiers trained on real data. This metric bridges the gap between waveform fidelity and clinical relevance, serving as a practical proxy for downstream utility. Our experiments show that over half of the synthetic samples are faithful, and filtering based on faithfulness improves classifier performance on arrhythmia detection tasks, particularly in low-resource settings. This suggests that faithfulness can be used both as an evaluation metric and as a selection strategy for curating synthetic datasets.

\paragraph{Insights from Ablation Studies.}  
Our modeling experiments highlight the diagnostic power of this benchmark. Removing mel-spectrogram supervision led to a 45\% increase in inter-lead correlation error and visibly distorted P and T waveforms, confirming the importance of frequency-domain losses for preserving global morphology. Similarly, ablations that removed demographic conditioning caused a regression toward population averages: QT intervals lost their age-appropriate scaling and personalization SNR dropped by 15\%. Privacy-aware training also proved critical—without it, membership inference risk rose significantly, indicating greater memorization of training samples. Together, these results demonstrate that each modeling component plays a complementary role in balancing fidelity, personalization, and privacy.

\paragraph{Error Analysis and Limitations.}  
Despite these advances, some limitations remain. Rare arrhythmias and edge-case morphologies (e.g., second-degree AV block, bundle branch blocks) remain underrepresented, likely reflecting class imbalance in PTB-XL. Demographic conditioning occasionally produced implausible combinations, such as exaggerated QRS duration for young patients, pointing to the need for more robust representation learning or explicit physiological constraints. Additionally, our evaluation was performed primarily on PTB-XL; external validation on MIMIC-IV or Chapman datasets would provide a stronger assessment of generalizability across acquisition settings and patient populations.

\paragraph{Privacy Evaluation as a Core Metric.}  
Unlike prior work, our benchmark explicitly incorporates privacy risk assessment using Membership Inference Risk (MIR) and Nearest Neighbor Adversarial Accuracy (NNAA). Our findings show that diffusion-based models trained with mel-spectrogram loss exhibit lower memorization risk, suggesting that frequency-domain supervision may act as an implicit regularizer. While encouraging, these results stop short of formal guarantees; future work should explore integrating differential privacy (e.g., DP-SGD) to provide provable privacy bounds for deployment in regulated clinical environments.

\paragraph{Guidelines and Best Practices.}  
From these findings, we propose several recommendations for the development and evaluation of synthetic ECG models:
(1) adopt multi-dimensional evaluation frameworks that go beyond RMSE and include morphological, clinical, and privacy metrics;
(2) leverage faithfulness both as an evaluation metric and as a filtering mechanism to curate high-utility synthetic datasets;
(3) use demographic conditioning selectively, monitoring outlier behavior and clinical feature distributions to prevent unrealistic outputs;
(4) incorporate frequency-domain losses when morphological realism is a priority, as they significantly improve waveform coherence;
and (5) contextualize results with real-vs-real baselines to interpret whether synthetic performance is within physiologically acceptable bounds.

\paragraph{Implications and Future Work.}  
Our results suggest that synthetic ECGs generated with MIDT-ECG can serve as a reliable drop-in replacement for real data in pre-training, low-resource model bootstrapping, or federated learning pipelines. Future directions include (a) expanding demographic conditioning to richer patient profiles, (b) validating cross-dataset generalization, (c) performing clinician-in-the-loop evaluation, and (d) exploring privacy-preserving training with formal guarantees. Beyond ECG, the proposed benchmarking framework could generalize to other biomedical time series (EEG, PCG, glucose monitoring), advancing the development of safe, trustworthy generative AI across healthcare.

\section{Conclusion}
In this work, we introduced \textbf{MIDT-ECG}, a principled framework to generate high-fidelity, personalized and privacy-preserving synthetic ECG data. By enhancing a state-of-the-art diffusion model with demographic-aware conditioning and mel-spectrogram-based supervision, we achieved significant gains in morphological realism and physiological coherence with (4\%-8\% gain) and notably a 70\% reduction in interlead correlation error, while lowering memorization privacy risk. Our comprehensive benchmark, which integrates statistical, morphological, clinical, and privacy metrics, provides a robust and clinically grounded evaluation protocol that can guide future model development. Beyond advancing synthetic ECG generation, our results highlight the broader importance of frequency domain supervision and faithfulness-based evaluation as tools for producing reliable biomedical time series data. Together, these contributions establish a scalable foundation for generating synthetic datasets that can bootstrap machine learning models, support federated learning, and enable privacy-preserving data sharing in healthcare. Future work will focus on extending conditioning to richer patient profiles, validating across multiple datasets, and providing formal privacy guarantees, which can bring us closer to effective generative AI systems for clinical research and decision support.

\newpage

\bibliography{sample}
\bibliographystyle{plain}

\clearpage
\appendix

\section{Outlier and Failure Mode Analysis}
\label{app:outliers}

In the main paper, we primarily report results conditioned on age and gender, as these attributes proved to be both effective and efficient. However, we also explored additional factors such as BMI and combinations of multiple attributes. For completeness, the following figures present results across all attribute types, providing a more comprehensive view of our conditioning framework.

\begin{figure}[h]  
\centering  
\includegraphics[width=\textwidth]{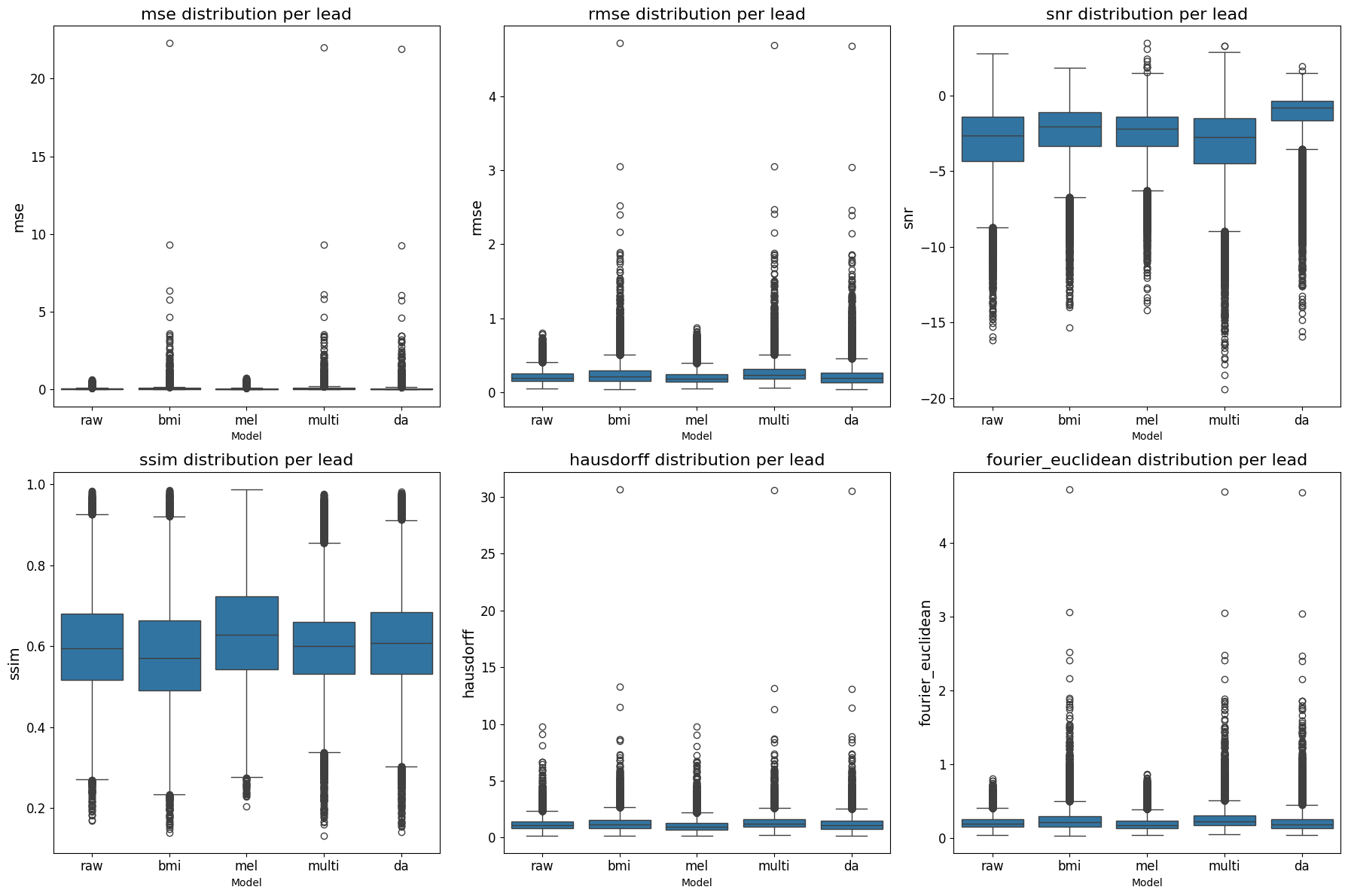}  
\caption{Statistical and Morphological metric distribution across baseline SSSD-ECG and 4 variants: mel - mel-spectrogram loss variant, multi - Disease + All demographic conditioned variant, bmi - Disease + BMI conditioned variant, da - Disease + Age conditioned variant. Boxplots show that conditioning models achieve higher SNRs compared to baseline, but exhibit a larger number of extreme outliers in error metrics (MSE, RMSE, Hausdorff distance, Fourier Transform distance), indicating greater variability and consistent occasional failure cases.}  
\label{fig:metric_disrtibution_appendix}  
\end{figure}

\begin{figure}  
\centering  
\includegraphics[width=\textwidth]{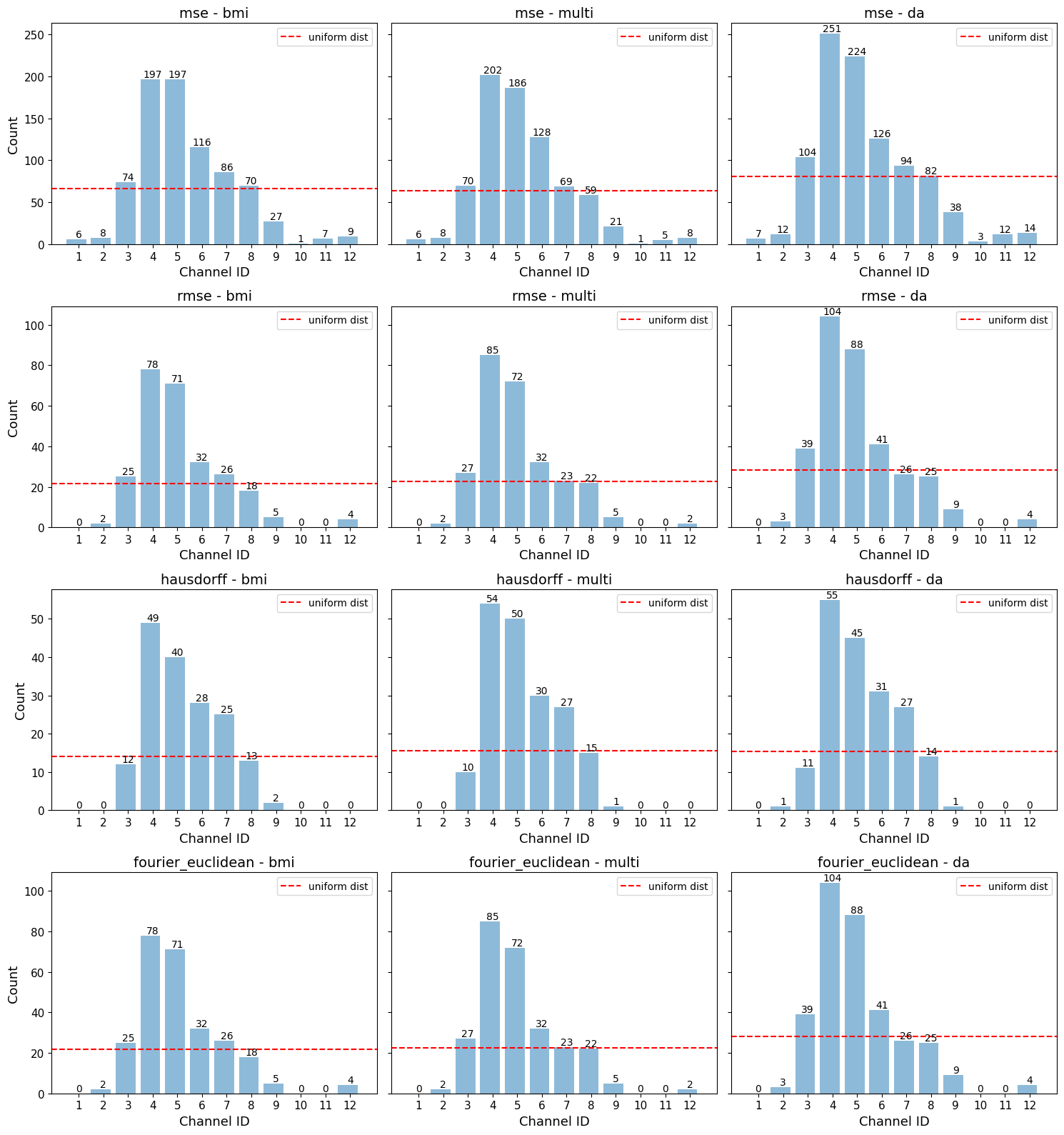}  
\caption{Lead distribution of outliers across conditioning variants in different metrics. Dashed red line represents the uniform distribution (evenly distributed across 12 leads). Lead 4 and 5 are observed to have consistent high frequency in outliers.}  
\label{fig:lead_distribution}  
\end{figure}

\begin{figure}  
\centering  
\includegraphics[width=\textwidth]{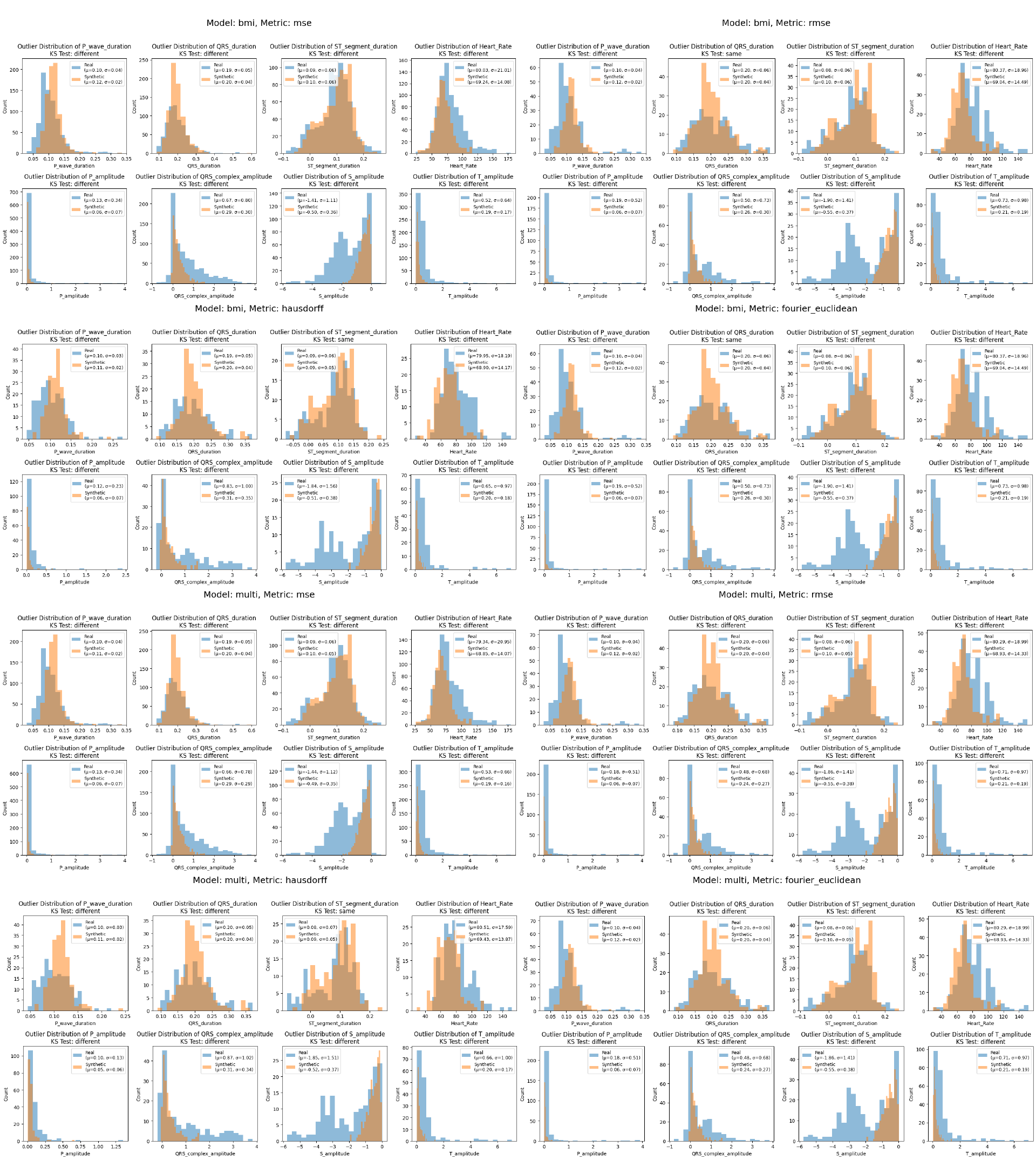}  
\caption{Clinical feature distribution for outlier synthetic ECGs generated by conditioned SSSD-ECG models (multi: D+all demographic, bmi: D+BMI, da: D+Age), identified based on MSE, RMSE, Hausdorff distance, and Fourier distance thresholds (Q3+3xIQR). Compared to the real ECG population,outliers from all conditioned models exhibit consistent deviations: lower heart rates, narrower QRS durations, and T-wave amplitudes. These trends suggest that while conditioning improves average signal quality, it may introduce systematic distortions in rare or complex cases, particularly impacting key clinical characteristics}  
\label{fig:clinical_distribution_1}  
\end{figure}

\begin{figure}  
\centering  
\includegraphics[width=\textwidth]{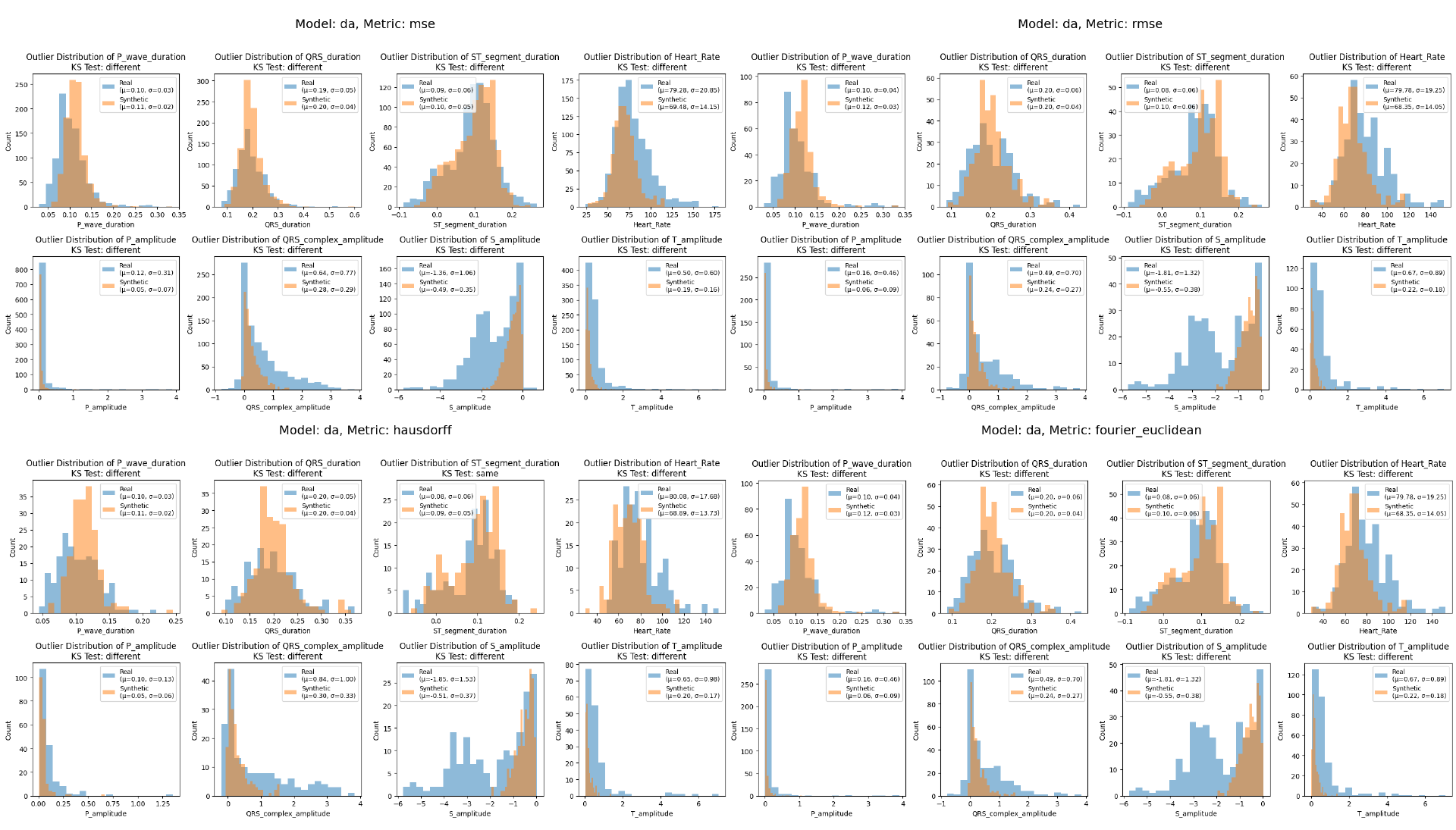}  
\caption{Clinical feature distribution for outlier synthetic ECGs generated by conditioned SSSD-ECG models - continued}  
\label{fig:clinical_distribution_2_appendix}  
\end{figure}

To better understand model limitations, we conducted an outlier analysis based on reconstruction error (RMSE). We found that demographically conditioned models tend to produce more extreme outliers, which disproportionately contribute to overall error. A clinical feature analysis of these outlier cases (Figures~\ref{fig:metric_disrtibution_appendix} to \ref{fig:clinical_distribution_2_appendix}) revealed that the models struggle most with atypical physiological states, such as bradycardia (low heart rate) and low-voltage ECGs. These cases are often under-represented in the training data and represent a key challenge for generative models, highlighting the importance of evaluating models not just on average performance but also on their robustness to rare events.


\section{Additional Visualizations}
This section provides supplementary visualizations that offer qualitative support for our quantitative findings and illustrate key aspects of our methodology and its practical application.

\subsubsection*{Qualitative Comparison of Real and Synthetic ECGs}
\begin{figure}  
\centering  
\includegraphics[width=\textwidth]{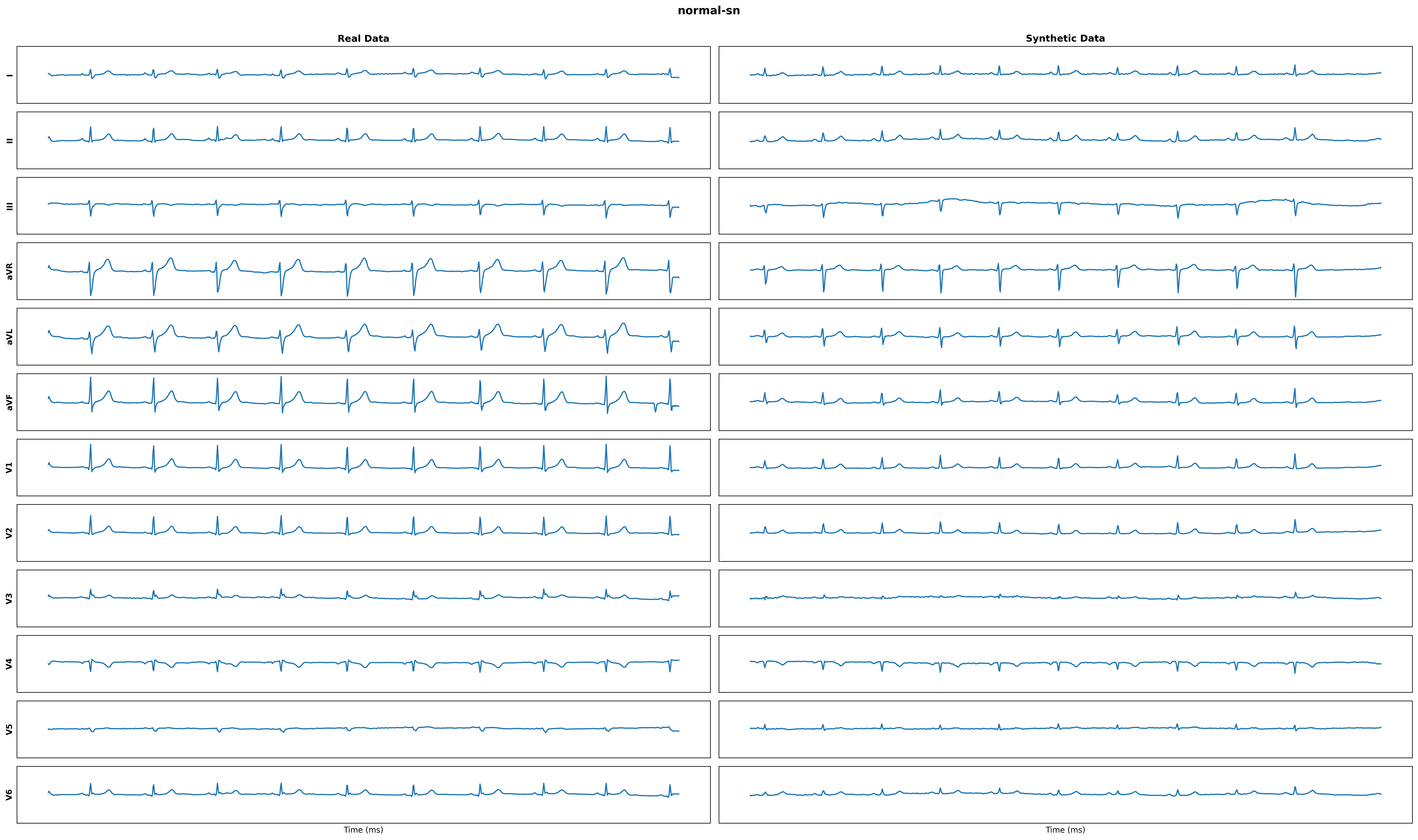}  
\caption{Comparison of real and synthetic 12-lead ECG signals for disease code 'norm-sn', with the synthetic sample generated by our MIDT model described in Table~\ref{tab:fidelity_and_privacy}.}  
\label{fig:ecgvisualization}  
\end{figure}

Figure~\ref{fig:ecgvisualization} provides a qualitative, side-by-side comparison of a real 12-lead ECG from the PTB-XL test set and a synthetic counterpart generated by our SSSD-ECG+A model for the same clinical condition ('norm-sn'). This visualization serves as a visual Turing test, demonstrating the model's ability to capture not only the fundamental P-QRS-T morphology and timing but also the subtle inter-lead relationships and overall rhythm characteristic of a real physiological signal. The high degree of visual similarity provides qualitative support for the strong quantitative performance reported in the main paper.

\subsubsection*{Illustrating the Mel-Spectrogram Loss Mechanism}
\begin{figure}  
\centering  
\includegraphics[width=\textwidth]{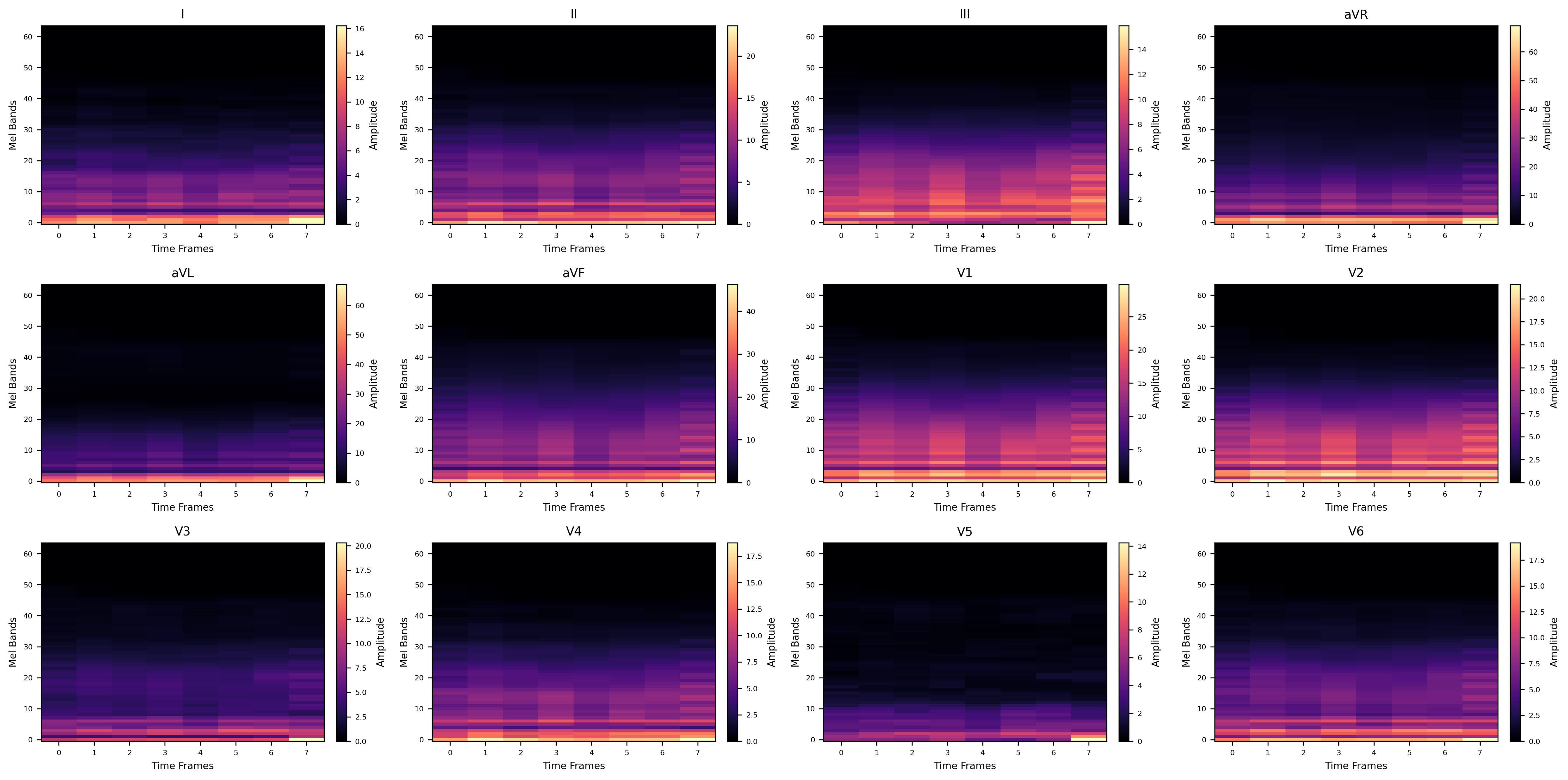}  
\caption{Mel-spectrogram visualization of the real 12-lead ECG signal (shown in Figure~\ref{fig:ecgvisualization}) after applying the Short-Time Fourier Transform (STFT)}  
\label{fig:ecgspectrogramvisualizationreal}  
\end{figure}

\begin{figure}  
\centering  
\includegraphics[width=\textwidth]{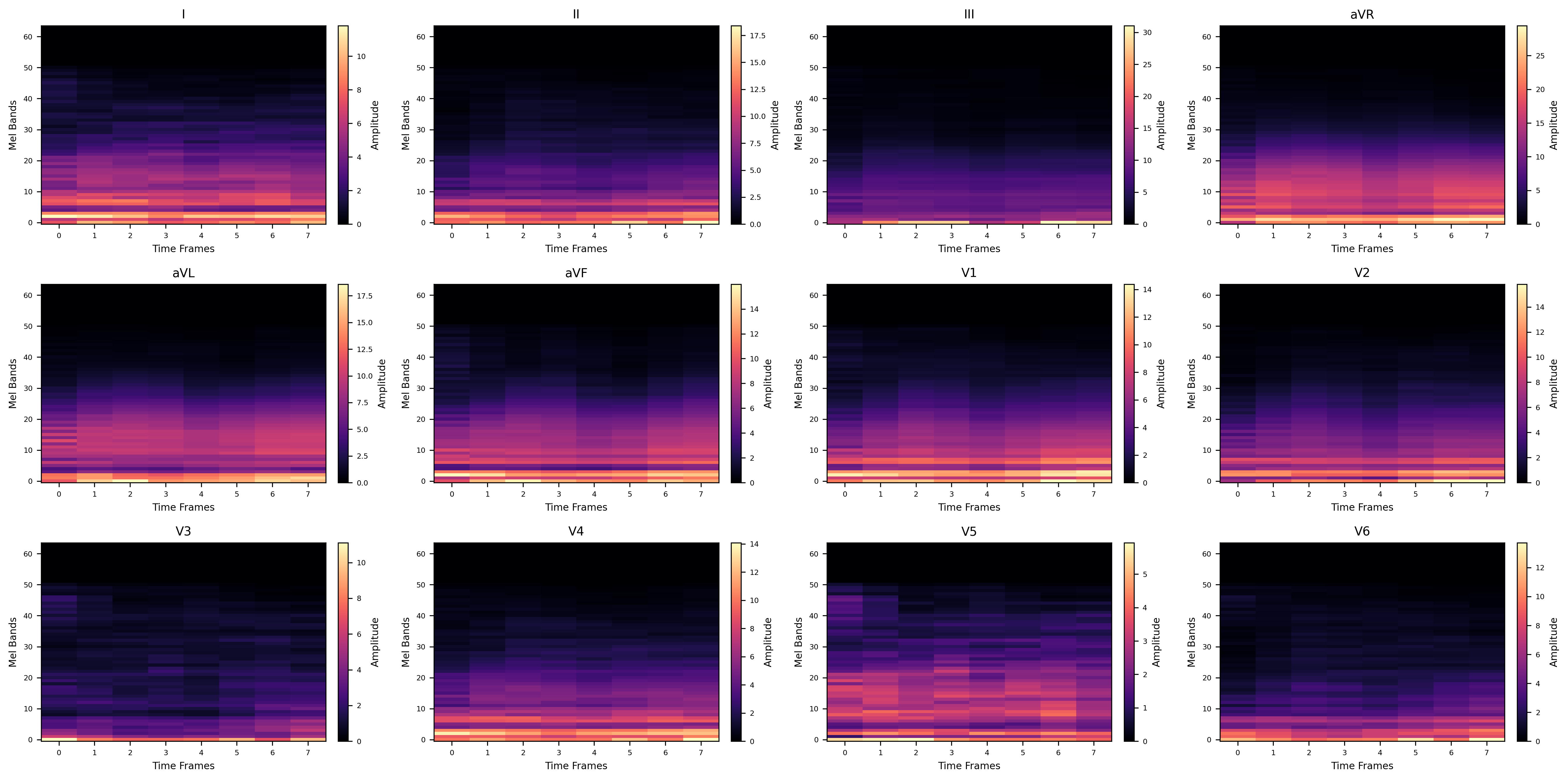}  
\caption{Mel-spectrogram visualization of the synthetic 12-lead ECG signal for disease code 'norm-sn' (shown in Figure~\ref{fig:ecgvisualization}) after applying the Short-Time Fourier Transform (STFT).}  
\label{fig:ecgspectrogramvisualizationsynth}  
\end{figure}

Figures~\ref{fig:ecgspectrogramvisualizationreal} and \ref{fig:ecgspectrogramvisualizationsynth} illustrate the core mechanism behind our mel-spectrogram loss function. They display the time-frequency representations (mel-spectrograms) of the real and synthetic ECGs shown in Figure~\ref{fig:ecgvisualization}, respectively. The loss function works by minimizing the pixel-wise difference between these two representations during training. The visual congruence between the two spectrograms—in terms of energy distribution across frequency bands and consistent temporal patterns—highlights how this frequency-domain supervision guides the model to reproduce the complex structural characteristics of the original signal. This directly leads to the improved morphological fidelity reported in our results.

\end{document}